\documentclass{ecai} 

\usepackage{latexsym}
\usepackage{amssymb}
\usepackage{amsmath}
\usepackage{amsthm}
\usepackage{booktabs}
\usepackage{enumitem}
\usepackage{graphicx}
\usepackage{color}
\usepackage{algorithm}
\usepackage{algorithmic}
\usepackage[switch]{lineno}
\usepackage{nicematrix}
\usepackage{subcaption}
\usepackage{multirow}
\usepackage{float}

\newtheorem{definition}{Definition}

\newif\ifshowcomments
\showcommentstrue
\ifshowcomments
\newcommand{\mynote}[2]{\textcolor{blue}{\fbox{\bfseries\sffamily\scriptsize#1}}
  \textcolor{blue}{{$/*$\textsf{\emph{#2}}$*/$}}}
\ifshowcomments

\ifshowcomments

\ifshowcomments

\else
\newcommand{\mynote}[2]{}
\fi

\usepackage{textcomp}
\usepackage{xcolor}

\newcommand{\BibTeX}{B\kern-.05em{\sc i\kern-.025em b}\kern-.08em\TeX}

\begin{document}

%%%%%%%%%%%%%%%%%%%%%%%%%%%%%%%%%%%%%%%%%%%%%%%%%%%%%%%%%%%%%%%%%%%%%%%%

%%% Use this command to specify your submission number.
%%% In doubleblind mode, it will be printed on the first page.
\begin{frontmatter}
%\paperid{470} 

%%% Use this command to specify the title of your paper.

\title{A Comprehensive Sustainable Framework for Machine Learning and Artificial Intelligence}

\author[A]{\fnms{Roberto}~\snm{Pagliari}}
\author[A]{\fnms{Peter}~\snm{Hill}}
\author[A,B]{\fnms{Po-Yu}~\snm{Chen} \thanks{Corresponding Author. Email: po-yu.chen11@imperial.ac.uk.}}
\author[A]{\fnms{Maciej}~\snm{Dabrowny}}
\author[A]{\fnms{Tingsheng}~\snm{Tan}}
\author[A]{\fnms{Francois}~\snm{Buet-Golfouse}}

\address[A]{JPMorgan}
\address[B]{Imperial College London}

\begin{abstract}
In financial applications, regulations or best practices often lead to specific requirements in machine learning relating to four key pillars: fairness, privacy, interpretability and greenhouse gas emissions. These all sit in the broader context of sustainability in AI, an emerging practical AI topic. However, although these pillars have been individually addressed by past literature, none of these works have considered all the pillars.
There are inherent trade-offs between each of the pillars (for example, accuracy vs fairness or accuracy vs privacy), making it even more important to consider them together. This paper outlines a new framework for Sustainable Machine Learning and proposes FPIG, a general AI pipeline that allows for these critical topics to be considered simultaneously to learn the trade-offs between the pillars better. 
Based on the FPIG framework, we propose a meta-learning algorithm to estimate the four key pillars given a dataset summary, model architecture, and hyperparameters before model training. This algorithm allows users to select the optimal model architecture for a given dataset and a given set of user requirements on the pillars. We illustrate the trade-offs under the FPIG model on three classical datasets and demonstrate the meta-learning approach with an example of real-world datasets and models with different interpretability, showcasing how it can aid model selection.
\end{abstract}

\end{frontmatter}

\section{Introduction}
\label{sec:intro}
Artificial Intelligence has become an emerging tool essential for all financial sectors \cite{hilpisch2020artificial,weber2023applications,pallathadka2023applications,umamaheswari2023role}.

However, the characterisation of AI extends beyond the realm of technology and permeates into the precincts of infrastructure \cite{crawford2021atlas} and ideology \cite{lanier_ai_2015}, leading to an opacity around the concept of AI \cite{katz_artificial_2020}. 
This nebulous nature of AI magnifies the challenges of effectively understanding and governing it while underscoring the need for malleability and interdisciplinary dialogue in AI ethics and governance. 
Consequently, this discourse does not gravitate towards a rigid definition of AI; rather, it embraces its polysemous essence and explores AI as a complex system \cite{bossel_modeling_2018}.

The current landscape of AI ethics frameworks \cite{fjeld_principled_2020,jobin_global_2019} is peppered with a proliferation of proposed principles and a conspicuous absence of uniformity across these frameworks. 
The initial environmental rights and climate justice movements were driven by the United Nations Climate Change Conferences,
and Sustainable Development Goals (SDGs) \cite{SDGs}, along with the environmental, social and corporate governance (ESG)
frameworks \cite{ESG}.
Unfortunately, while the environmental implications of AI are gradually entering the discourse \cite{strubell_energy_2019}, the broader concept of sustainability in AI appears to be largely overlooked \cite{hagendorff2022blind}.
Recent literature only fostered a narrow vision of sustainable AI \cite{van2021sustainable,wu2018learning}, neglecting the interconnected nature of various AI governance challenges. 
A holistic view of sustainable AI should be an amalgamation of three intertwined pillars: economic, environmental and social, necessitating a complex systems approach \cite{de2023sustainability}.

% Finance Motivation
Financial institutions have particular duties in relation to AI that need to be paid close attention to. The Information Commissioner's Office (ICO) has strict guidance on AI regarding interpretability, data protection and privacy \cite{ico}. Additionally, there have been several recent developments from significant organisations relating to AI regulations, bolstering the importance of sustainable AI's key features. For example, the European Union has proposed the AI Act \cite{aiact}, a European law on AI. The Bank of England has also recently updated their model risk management framework \cite{BOE}, outlining the expectations of the Prudential Regulation Authority (PRA) for banks' management of model risk. They indicate the need for a robust approach to model risk and discuss five key principles within their framework, from governance to validation. The particular focus on model risk mitigants indicates the importance for banks to consider factors such as the interpretability and fairness of their models as part of the model selection stage. In addition, the FCA has also recently updated their consumer duty expectations \cite{ConsumerDuty}, raising the standards required by financial institutions from previous expectations. With the growing role AI is having, there is an increased expectation from the FCA for consumers to be at the forefront of model design. The EU AI act \cite{EUAI} has also been proposed recently, focusing on the risk of AI applications and categorising AI use into four risk levels, as well as imposing increased documentation and validation of models.

\begin{figure}[t!]
\centering
\includegraphics[width=0.4\textwidth]{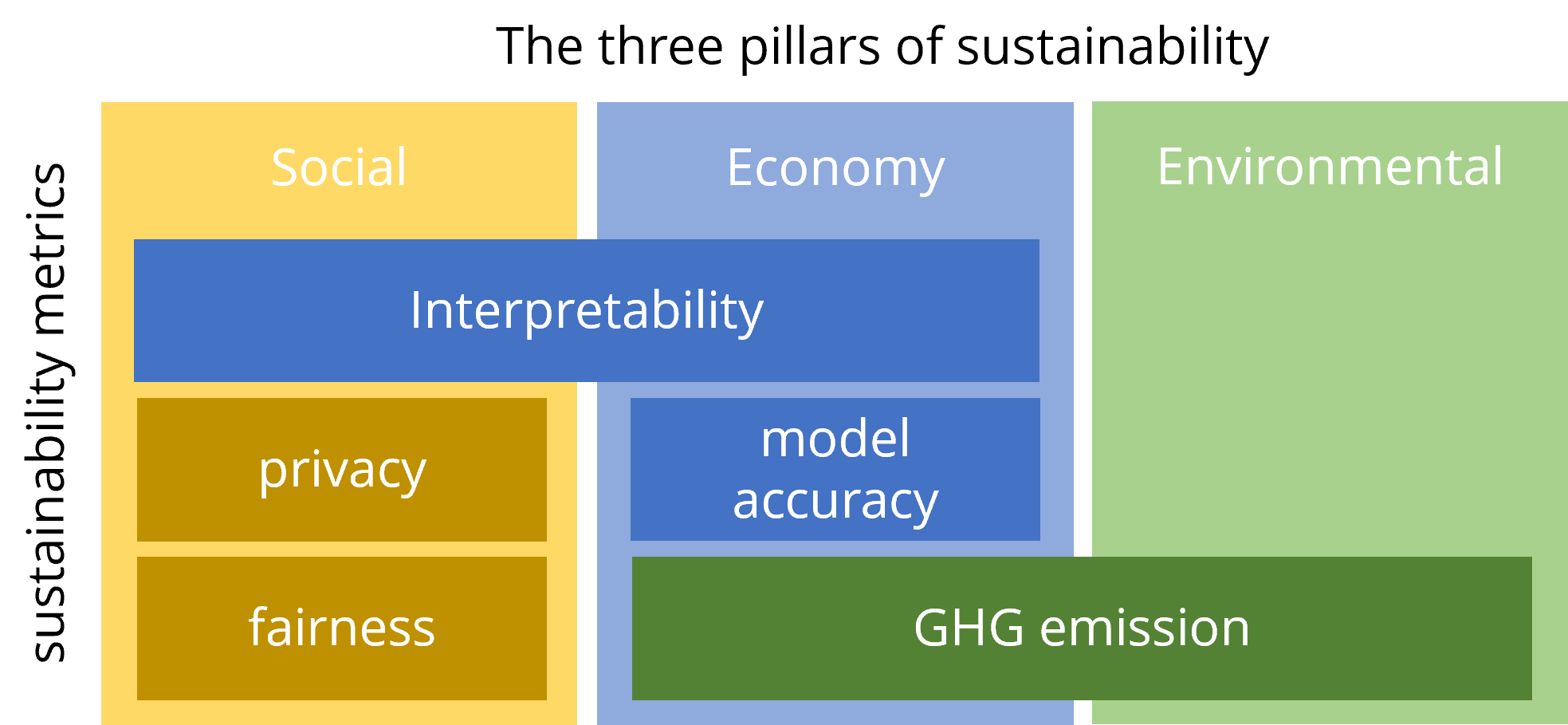}
\vspace{.2cm}
\caption{Privacy, fairness, accuracy, interpretability and GHG emission are the practical sustainability metrics that drive the three pillars of sustainability.}
\vspace{1cm}
\label{fig:concept}
\end{figure}

In this paper, we first advocate for the adoption of the principles of sustainability science to AI, analysing AI through the lens of an unsustainable system. We then propose a Fair, Private, Interpretable and Green (FPIG) framework to address the above-mentioned pillars. These four features (as illustrated in Fig \ref{fig:concept}) are tightly associated with the concept of sustainability but, to the best of our knowledge, have yet to be tackled together under a single framework.

In the FPIG framework, we first propose to integrate fairness \cite{hardt2016equality} into our ML objective function to reduce the loss disparity across groups. This approach allows us to change the level of fairness required in our training, varying from a standard optimisation (with no additional fairness constraints) to a multi-objective scenario, where trade-offs across different metrics form the Pareto front.
Secondly, we integrate the concept of differential privacy during the model training process. By adding noise during training, we ensure that good models are obtained for all other dimensions across varying degrees of differential privacy. 

The carbon dioxide (CO2) emission during model training and inference pipeline is tracked and monitored by an independent software package, CodeCarbon \cite{Codecarbon}. Ultimately, we propose a new meta-learning algorithm that helps users to find better AI models and hyperparameters (e.g., number of neural network layers) given a dataset and the three sustainability goals plus model interpretability.

To demonstrate the effectiveness of the FPIG framework, we evaluate it with five independent datasets and four distinctive types of machine learning (ML) models, varying in interpretability. We compute the results across the different pillars. Our evaluation reveals common trade-offs from training models using our frameworks, such as trade-offs between accuracy, fairness and privacy. 
We also indicate the significant features when considering a meta-learning approach, allowing us to estimate the impact on accuracy, fairness, privacy and carbon emissions of training different models without needing to undertake the cost of actually training them. We demonstrate the usefulness of this approach on a particular example.

% 8. PAPER ORGANISATION
The rest of this paper is organised as the following. A brief overview of the related work is given in Section \ref{sec:related_work}. We then present the FPIG framework in Section \ref{sec:framework}. The experiment results with five distinctive datasets are presented in Section \ref{sec:evaluation}, and the paper is ultimately concluded in Section \ref{sec:conclusion}.

\section{Sustainability in Artificial Intelligence and Machine Learning}
\label{sec:related_work}

The notion of "sustainable AI" has been proposed by a variety of researchers and practitioners \cite{coeckelbergh_ai_2021,saetra2021ai} intending to emphasise the interconnection between AI and sustainability \cite{van2021sustainable}. 
Nevertheless, the term "sustainable" is frequently interpreted as synonymous with "environmentally friendly." 
For instance, the "Sustainable AI" manifesto issued by Facebook AI \cite{wu2018learning} is solely focused on diminishing carbon emissions from AI systems whilst vowing to "advance the field of AI in an environmentally responsible manner". 
This exemplifies the challenge of encouraging stakeholders to embrace a multifaceted perspective on sustainability rather than confining it merely to environmental aspects.

%
% Fairness
\paragraph{Fairness} is one of the most important sustainability metrics according to SDGs.
Fairness in ML models refers to the absence of bias or discrimination in the predictions and decisions made by the models \cite{narayanan2018translation,Verma2018,richardcompas}. 
Technically speaking, fairness involves identifying and addressing biases in the data used to train the models and the algorithms themselves. 
Techniques such as pre-processing the data to remove biased patterns, using specialised algorithms that explicitly consider fairness constraints during training, and employing fairness metrics to evaluate model performance can help achieve fairness in ML models \cite{buet2022towards,buet2023fairness}.
Recent research has defined different fairness metrics for AI \cite{narayanan2018translation,Verma2018,richardcompas}. Among these definitions, group fairness metrics such as demographic parity \cite{Calders2010}, equalised odds \cite{hardt2016equality}, and social fairness ensures that different groups based on a protected attribute are treated equally. The impossibility theorems on fairness \cite{kleinberg2016inherent,chouldechova2017fair} show that these definitions cannot all be satisfied at once. There is also individual fairness and counterfactual fairness \cite{kusner2017counterfactual}, which are proposed to ensure fairness at individual levels. Fairness can be considered in both a supervised learning or an unsupervised learning setting. Recent research also investigated in incorporating multiple fairness objectives into ML models given a desired level of fairness, using group functionals \cite{buet2022towards,buet2023fairness}.  

\begin{table}[t!]
    \centering
    \resizebox{0.48\textwidth}{!}
    {
    \begin{NiceTabular}{|c|c|c|}
        \hline 
        \textbf{Model} & \textbf{Explainability} & \textbf{Tunable Hyperparameters} \tabularnewline
        \hline \hline
        Linear Regression  & {1} & { Regularisation strength} \tabularnewline
        \hline 
        Tree  & {1}  & {Max depth} \tabularnewline
        \hline
        \multirow{3}{*}{Random Forest}  & \multirow{3}{*}{2} & {Number of estimators}  \\
        & & {Max depth} \\
        & & {Max rows to subsample}
        \tabularnewline
        \hline
        \multirow{9}{*}{XGBoost}  & \multirow{9}{*}{2} & {Number of estimators} \\
        & & {Max depth} \\
        
        & & {Learning rate} \\

        & & {Fraction of columns to subsample} \\

        & & {Max rows to subsample} \\

        & & {L1 regularisation} \\
        & & {L2 regularisation} \\
        & & {Minimum loss reduction for partition} \\
        & & {Balance between positive and negative samples} 
        \tabularnewline
        \hline
         \multirow{2}{*}{Neural Network}  & \multirow{2}{*}{3} & {Number of layers} \\
        & & {Layer size} \\
        \hline
    \end{NiceTabular}
    }
    \vspace{0.3cm}
    \caption{Models and associated tunable hyperparameters used in our benchmarking study. Note that we consider a subset of the hyperparameters that could potentially be tuned, and thus this is not an exhaustive list. A self-defined measure of explainability from $1$ to $3$ is included, with $1$ being the most and $3$ least explainable.}
    \vspace{0.3cm}
    \label{tab:datasets_2}
\end{table}

% Privacy
\textbf{Privacy} in ML models pertains to preserving the confidentiality and security of sensitive data used for training and inference \cite{solove2008understanding}. Privacy protection involves implementing mechanisms that prevent unauthorised access, use, or disclosure of personal information.
Techniques like data anonymisation, encryption, and secure multi-party computation can be employed to protect privacy in ML models.
Differential privacy (DP) has also been regarded as the gold standard in academia as it provides a well-defined theoretical guarantee. It can be applied to ensure that individual data points are not distinguishable, thereby safeguarding privacy while maintaining the utility of the models \cite{abadi2016deep}.
Adding noise during training is one way to incorporate DP into ML models. For example, Differentially private stochastic gradient descent (DP-SGD) \cite{dwork2014algorithmic} makes deep learning models differentially private by modifying the mini-batch stochastic optimisation process during gradient descent \cite{chen2022dpgen, chen2020gs, horigome2023local}. Other approaches incorporate DP to data synthesis \cite{lin2022dpview, vietri2022private} and than train the model on the DP synthetic data, making these models more robust against DP attacks with exponentially many queries.

% Time series figures tracking model performance along training.
\begin{figure*}[t!]
\centering
    % Fig (a)
    \begin{subfigure}{0.33\textwidth}
        \includegraphics[width=\linewidth]{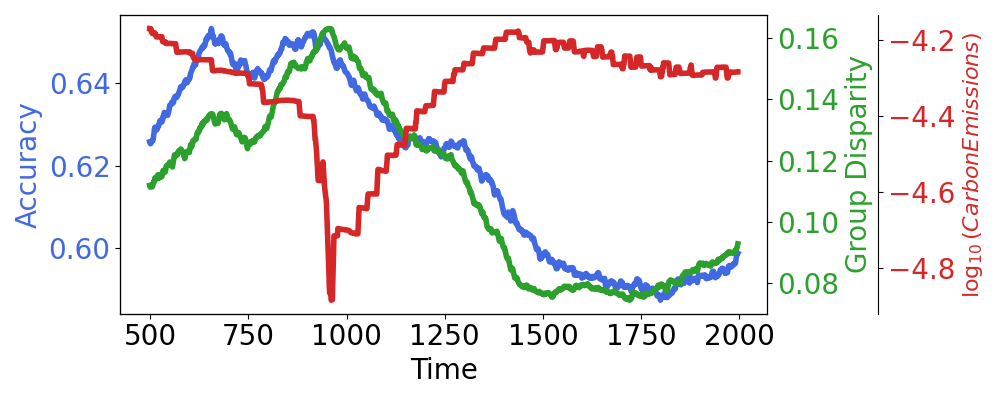}
        \vspace{-15pt}\caption{Adult Income}\vspace{9pt}
    \end{subfigure}
    % Fig (b)
    \begin{subfigure}{0.33\textwidth}
        \includegraphics[width=\linewidth]{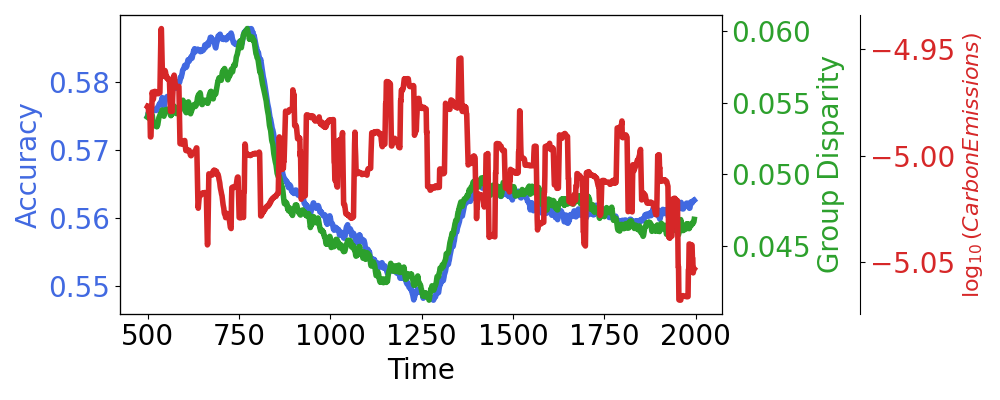}
        \vspace{-15pt}\caption{COMPAS}\vspace{9pt}
    \end{subfigure}
    % Fig (c)
    \medskip
    \begin{subfigure}{0.33\textwidth}
        \includegraphics[width=\linewidth]{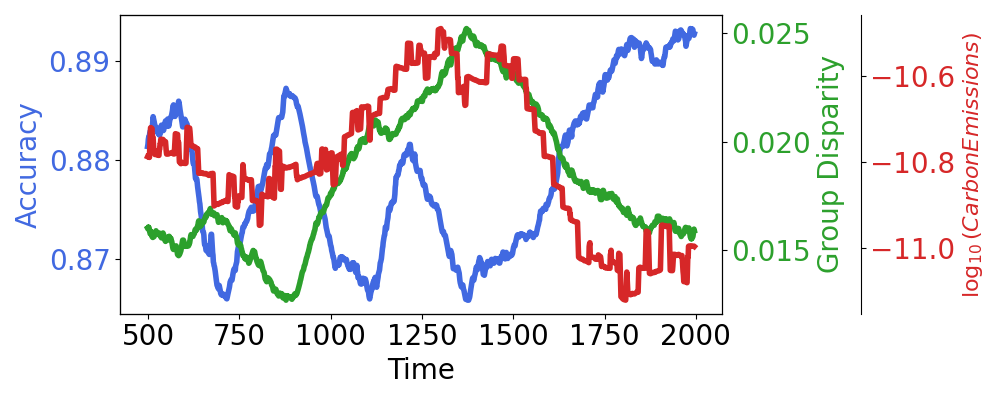}
        \vspace{-15pt}\caption{LSAC}\vspace{9pt}
    \end{subfigure}
   
    % Fig (d)
    \begin{subfigure}{0.33\textwidth}
        \includegraphics[width=\linewidth]{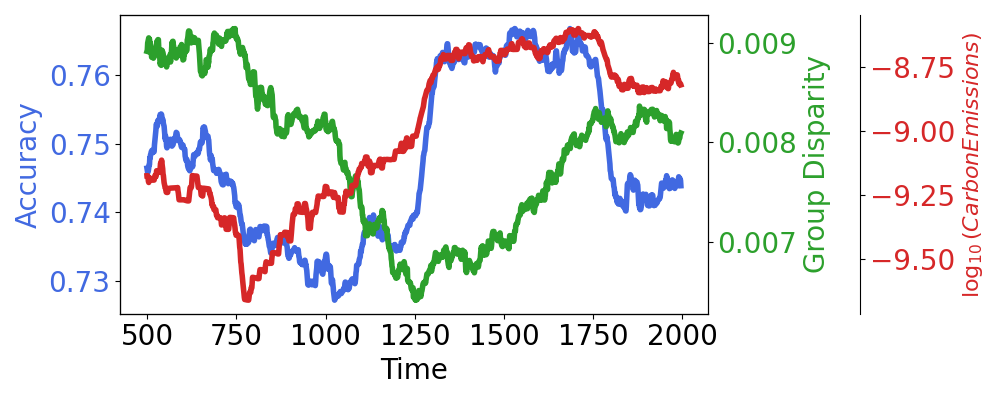}
        \vspace{-15pt}\caption{Loan Default}\vspace{6pt}
    \end{subfigure}
    % Fig (e)
    \medskip
    \begin{subfigure}{0.33\textwidth}
        \includegraphics[width=\linewidth]{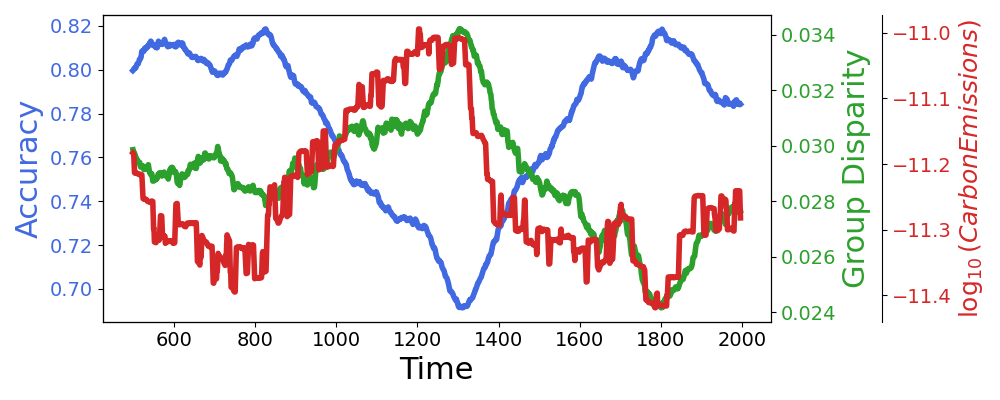}
        \vspace{-15pt}\caption{Support 2}\vspace{6pt}
    \end{subfigure}
    \vspace{3pt}
\caption{The moving average (we consider a rolling $500$ trial period.) of Accuracy, Group Disparity, and Emissions metrics over time (number of trials), for (a) Adult income, (b) COMPAS, (c) LSAC, (d) Loan Default, (e) Support2 datasets.}
\vspace{6pt}
\label{fig:optuna_trials}
\end{figure*}

%
% Interpretability
\textbf{Interpretability} in ML models refers to understanding and explaining the reasoning behind the model's predictions or decisions \cite{novakovsky2023obtaining}. Interpretability techniques involve feature importance analysis, rule-based approaches, and model-agnostic techniques\cite{dwivedi2023explainable}. These techniques provide insights into the factors influencing the model's output and enable humans to comprehend and validate the decision-making process. Also, layer-wise relevance propagation (LRP) or attention mechanisms can help identify relevant features or parts of the input contributing to the model's predictions.
A few works investigated the trade-offs between computational efficiency and model explainability \cite{graziani2023global, lin2022model}.

%
% Greenhouse Gas
\textbf{Greenhouse Gas (GHG) emissions} associated with ML models refer to the carbon footprint generated during the model's lifecycle, including data processing, training, inference, and deployment \cite{henderson2020towards}. 
At a technical level, reducing GHG emissions involves optimising the computational resources used for training by employing energy-efficient hardware and algorithms. 
Techniques like model compression, which reduces the model's size or complexity, can also contribute to lower energy consumption during inference \cite{choudhary2020comprehensive}. 
Additionally, adopting hardware acceleration techniques, using distributed computing, and leveraging renewable energy sources can help minimise the environmental impact of ML models \cite{lai2020survey}.

% Carbon Emission and Model Interpretability
While there exists extensive literature working on improving model efficiency, which ultimately leads to lower greenhouse gas (GHG) emissions, explainable AI (xAI) has become an emerging area that attracts research \cite{dwivedi2023explainable, novakovsky2023obtaining}. A few works investigated the trade-offs between computational efficiency and model explainability \cite{graziani2023global, lin2022model}. Nevertheless, to our knowledge, this paper is the first to introduce a framework encompassing all four sustainability goals, including fairness, privacy, model interpretability and low GHG emissions.

\section{The FPIG Framework}
\label{sec:framework}

We propose a framework incorporating the four sustainability features into the model training pipeline. Specifically, we are optimizing along different dimensions (e.g., model performance, explainability, carbon emission and fairness with some level of privacy).

\subsection{Single-Objective Optimization}
Traditionally, the hyper-parameters of a machine learning model are tuned to maximize one metric of interest, for example, the area under the curve. Once the metric of interest is defined, the objective is to minimize the quantity in Eq. \eqref{eq:single_obj}:

\begin{equation}\label{eq:single_obj}
\begin{aligned}
&\text{minimize}    & & f(\textbf{x}) \\
&\text{subject to}  & & \textbf{x} \in \mathbf{X} 
\end{aligned}
\end{equation}
where $\textbf{x} \in \mathbf{X} \subseteq \mathbb{R}^d$ is the set of $d$ hyper-parameters, $\mathbf{X} \subseteq \mathbb{R}^d$ is the search space and $f(\cdot)$ is the objective function, for example, a loss function to be minimized. 

In the case of a single-objective scenario, such as the one shown in Eq. \eqref{eq:single_obj}, the Tree-structured Parzen Estimator (TPE), originally proposed for neural networks \cite{bergstra2011algorithms}, is widely used for optimizing a wide number of machine learning models. The key idea is to separate the likelihood function $p(\mathbf{x}|y)$ in two components so as to identify which region the best hyper-parameters are likely to be in: 
\begin{equation}\label{eq:tpe}
  p(\mathbf{x}|y) =
    \begin{cases}
      l(\mathbf{x}) & \text{if } y < y^*  \\
      g(\mathbf{x}) & \text{if } y \geq y^*
    \end{cases}       
\end{equation}
where $y^*$ is usually a quantile of the observed values $y$ (e.g, $80\%$), and  $l(\cdot)$ and $g(\cdot)$ are the probability density functions formed using the observations $\{\mathbf{x}^{(i)}\}$ below and above $y^*$, respectively. This methodology begins with several random observations $\{\mathbf{x}\}^{(i)}$ and proceeds iteratively by adding one observation at a time such that the expected improvement is maximized. As shown in \cite{bergstra2011algorithms}, the expected improvement is proportional to 
\begin{equation*}
\text{EI}_{y^*}(\mathbf{x}) \propto \left[ y^* + \frac{g(\mathbf{x})} {l(\mathbf{x})} (1-y^*) \right]^{-1}.
\end{equation*}

In other words, the aim is to sample with higher probability under $l(\mathbf{x})$ (i.e., the portion of the density with the most promising hyper-parameters) and lower probability under $g(\mathbf{x})$.

\subsection{Multi-Objective Optimization}

In a multi-objective scenario, we are interested in the minimization (or maximization) of many objectives that usually conflict. The optimization problem is defined as follows:  
\begin{equation}\label{eq:multi_obj}
\begin{aligned}
&\text{minimize}    & & f(\mathbf{x}) = (f_1(\textbf{x}), \dots, f_n(\textbf{x})) \\
&\text{subject to}  & & \textbf{x} \in \mathbf{X} 
\end{aligned}
\end{equation}
where, as in Eq. \eqref{eq:single_obj}, $\textbf{x} \in \mathbf{X} \subseteq \mathbb{R}^d$ is the set of $d$ hyper-parameters, $\mathbf{X} \subseteq \mathbb{R}^d$ is the search space. The difference is that this time, we define a vector of cutoffs $\mathbf{Y}^* = (y_1^*,\dots, y_n^*)$, that is, a cutoff for every objective, and the TPE estimator is generalized a follows: 
\begin{equation}\label{eq:tpe_mo}
  p(\mathbf{x}|\mathbf{y}) =
    \begin{cases}
      l(\mathbf{x}) & \text{if } \mathbf{y} \succ \mathbf{Y^*} \cup \mathbf{y} \parallel \mathbf{Y^*}  \\
      g(\mathbf{x}) & \text{if } \mathbf{Y^*} \succeq \mathbf{y}
    \end{cases}       
\end{equation}
where the $\succ$, $\succeq$ and $\parallel$ operators denote dominant, weakly dominant and non-comparable relationships, respectively, as per \cite{ozaki2020multiobjective}. This time, as shown in Eq. \eqref{eq:tpe_mo}, the most "promising" solutions are those that dominate the cutoff 
$\mathbf{Y^*}$ or those that are not comparable to $\mathbf{Y^*}$. The less promising models are those that are weakly dominated by $\mathbf{Y^*}$ and, thus, contribute to forming the density function of the least "promising" hyper-parameters $g(\mathbf{x})$. Splitting the data into two sets is, in this instance, achieved via the Hype method \cite{bader2011hype}; however, the methodology is, in principle, the multi-objective equivalent of Eq. \eqref{eq:single_obj}.

%% TABLE
%
\begin{table*}[ht!]
    \centering
    \resizebox{\textwidth}{!}
    {
    \begin{NiceTabular}{|c|c|c|c|c|c|c|c|}
        \hline  
        \textbf{Dataset} & \textbf{Best Model w.r.t.} & \textbf{Model Architecture} & \textbf{Accuracy} & \textbf{Group Disparity} & \textbf{Differential Privacy} & \textbf{Explainability} & \textbf{Carbon Emissions} \tabularnewline \hline \hline
        %
        % ADULT
        \multirow{4}{*}{Adult income \cite{misc_adult_2}} & Accuracy &  xgboost & \textbf{0.861} & 0.167 & 10.5 & 2.0 & $3.92 \times 10^{-6}$
        \tabularnewline \cline {2-8}
        & Fairness & xgboost & 0.767 & \textbf{0.000} & 10.0 & 2.0 & $3.18 \times 10^{-6}$ \tabularnewline
        \cline {2-8}
        & Carbon Emissions & decision tree & 0.726 & 0.446 & 10.5 & 1.0 & $\mathbf{2.26 \times 10^{-6}}$ \tabularnewline
        \cline {2-8}
        & Equal Importance & xgboost & 0.767 & 0.000 & 10.0 & 2.0 & $3.05 \times 10^{-6} $ \tabularnewline \hline \hline
        %
        % COMPAS
        \multirow{4}{*} {COMPAS \cite{compas_2}}  & Accuracy & decision tree & \textbf{0.671} & 0.095 & 10.5 & 1.0 & $1.65 \times 10^{-7}$
        \tabularnewline \cline {2-8}
        &   Fairness & xgboost & 0.460 & \textbf{0.000} & 0.5 & 2.0 & $6.59 \times 10^{-7}$\tabularnewline \cline {2-8}
        &  Carbon Emissions & decision tree & 0.630 & 0.079 & 1.5 & 1.0 & $\mathbf{7.00 \times 10^{-8}}$ 
        \tabularnewline
        \cline {2-8}
        &  Equal Importance & logistic regression & 0.614 & 0.007 & 0.5 & 1.0 & $5.02 \times 10^{-7}$ \tabularnewline
        \hline \hline
        %
        % LSAC
        \multirow{4}{*}{LSAC \cite{lsac}} & Accuracy  & neural network & \textbf{0.949} & 0.003 & 10.5 & 3.0 & $9.71 \times 10^{-5}$ 
        \tabularnewline \cline {2-8}
        & Fairness & xgboost & 0.946 & \textbf{0.000} & 2.0 & 2.0 & $7.58 \times 10^{-7}$ \tabularnewline
        \cline {2-8}
        & Carbon Emissions & xgboost & 0.946 & 0.000 & 2.0 & 2.0 & $\mathbf{7.58 \times 10^{-7}}$\tabularnewline
        \cline {2-8}
        & Equal Importance & xgboost & 0.946 & 0.000 & 2.0 & 2.0 & $7.58 \times 10^{-7}$ \tabularnewline \hline \hline
        %
        % LOAN Default
        \multirow{4}{*}{Loan Default \cite{loandataset}} & Accuracy & neural network & \textbf{0.999} & 0.013 & 10.5 & 3.0 & $2.56 \times 10^{-4}$\tabularnewline
        \cline {2-8}
        & Fairness & decision tree & 0.745 & \textbf{0.000} & 10.5 & 1.0 & $3.59 \times 10^{-6}$\tabularnewline
        \cline {2-8}
        & Carbon Emissions & decision tree & 0.745 & 0.000 & 10.5 & 1.0 & $\mathbf{3.59 \times 10^{-6}}$\tabularnewline
        \cline {2-8}
        & Equal Importance & random forest & 0.848 & 0.0003 & 10.5 & 2.0 & $1.44 \times 10^{-5}$ \tabularnewline \hline \hline
        %
        % SUPPORT 2
        \multirow{4}{*}{Support2 \cite{loandataset}} & Accuracy & xgboost & \textbf{0.977} & 0.0259 & 10.5 & 2.0 & $3.01 \times 10^{-6}$
        \tabularnewline 
        \cline {2-8}
        & Fairness & xgboost & 0.255 & \textbf{0.000} & 2.5 & 2.0 & $1.64 \times 10^{-6}$\tabularnewline
        \cline {2-8}
        & Carbon Emissions & decision tree & 0.913 & 0.021 & 10.5 & 1.0 & $\mathbf{6.16 \times 10^{-7}}$\tabularnewline
        \cline {2-8}
        & Equal Importance & decision tree & 0.956 & 0.005 & 10.5 & 1.0 & $7.02 \times 10^{-7}$\tabularnewline \hline
    \end{NiceTabular}
    }
    \vspace{0.5cm}
    \caption{The best models across different datasets, concerning different sustainability metrics. We consider accuracy, fairness, carbon emissions and an equal importance approach, as defined in Section \ref{subsec:pareto_front}. The results shown are on an out-of-sample test set.}
    \vspace{0.5cm}
    \label{tab:best_models}
\end{table*}

\subsection{Incorporating Fairness}
\label{subsec:fairness}
There exist several metrics for quantifying fairness and algorithmic bias of machine learning models. Amongst them, the most popular and often considered are equalized odds, equal opportunity, and demographic parity \cite{mehrabi2021survey}. Without loss of generality, in our framework, we optimized for demographic parity, which is satisfied when the condition below holds true:  
\begin{equation*}
    P(\hat{Y} | A=0) = P(\hat{Y} | A=1).
\end{equation*}
11121321In other words, the protected attribute $A$ (e.g., sex or age) does not influence the model's outcome. In reality, demographic parity can never be exactly zero because the protected attribute $A$ usually correlates with other features the model uses. Hence the objective is minimizing group disparity $f(\hat{Y}, A)$ as

\begin{equation}\label{eq:group_disparity}
    f(\hat{Y}, A) = \left | P(\hat{Y} | A=0) - P(\hat{Y} | A=1) \right |.
\end{equation}

\subsection{Incorporating Privacy}
To include privacy in our framework, we consider the concept of differential privacy \cite{dwork2006calibrating, dwork2014algorithmic}, which we use to include privacy guarantees in the model.

\begin{definition}
    A randomized mechanism $\mathcal{M}:\mathcal{D} \rightarrow \mathcal{R}$ satisfies $(\epsilon, \delta)$-differential privacy if for any two adjacent inputs $d,d' \in \mathcal{D}$, and any $S \subset \mathcal{R}$ fulfil the inequality below:
    \begin{equation}
    \mathbb{P}(\mathcal{M}(d) \in S) \leq e^\epsilon \mathbb{P}(\mathcal{M}(d') \in S) + \delta.
    \end{equation}
\end{definition}

To incorporate Differential Privacy into diverse model architectures, we didn't take the common route where privacy is applied via training, such as the popular Differentially-Private Stochastic Gradient Descent (DP-SGD) \cite{abadi2016deep}. Instead, we exploit the idea that converts data into differentially private synthetic data, which can be exploited by different model architectures universally. Various DP data synthesis approaches were proposed for different modalities. For example, \cite{vietri2022private} and \cite{lin2024differentially} can generate DP synthetic data for tabular and images, respectively. 

In this work, we adopted DPView \cite{lin2022dpview}, a state-of-the-art DP-aware high-dimensional data synthesis for tabular data. For a given privacy requirement $\epsilon$, it utilises the domain size of attributes and the correlation among attributes to analytically optimise both privacy budget allocation and consistency in producing synthetic data points. 
Their evaluation demonstrated that the approach is versatile (when applied to tabular data from vest applications) and can effectively preserve model utilities. 
Compared to traditional gradient-based approaches (e.g., DP-SGD \cite{abadi2016deep}) where privacy can still be breached by querying the model multiple times and the total privacy budgets are required to split across users, DPView is more robust as it does not suffer from the same issues since noises are directly applied to data instead of the model during training. 

%% Fig - Scatter Plots
%%
% Time series figures tracking model performance along training.
\begin{figure*}[ht!]
\centering
    % Fig (b)
    \begin{subfigure}{0.33\textwidth}
        \vspace{-12pt}\includegraphics[width=\linewidth]{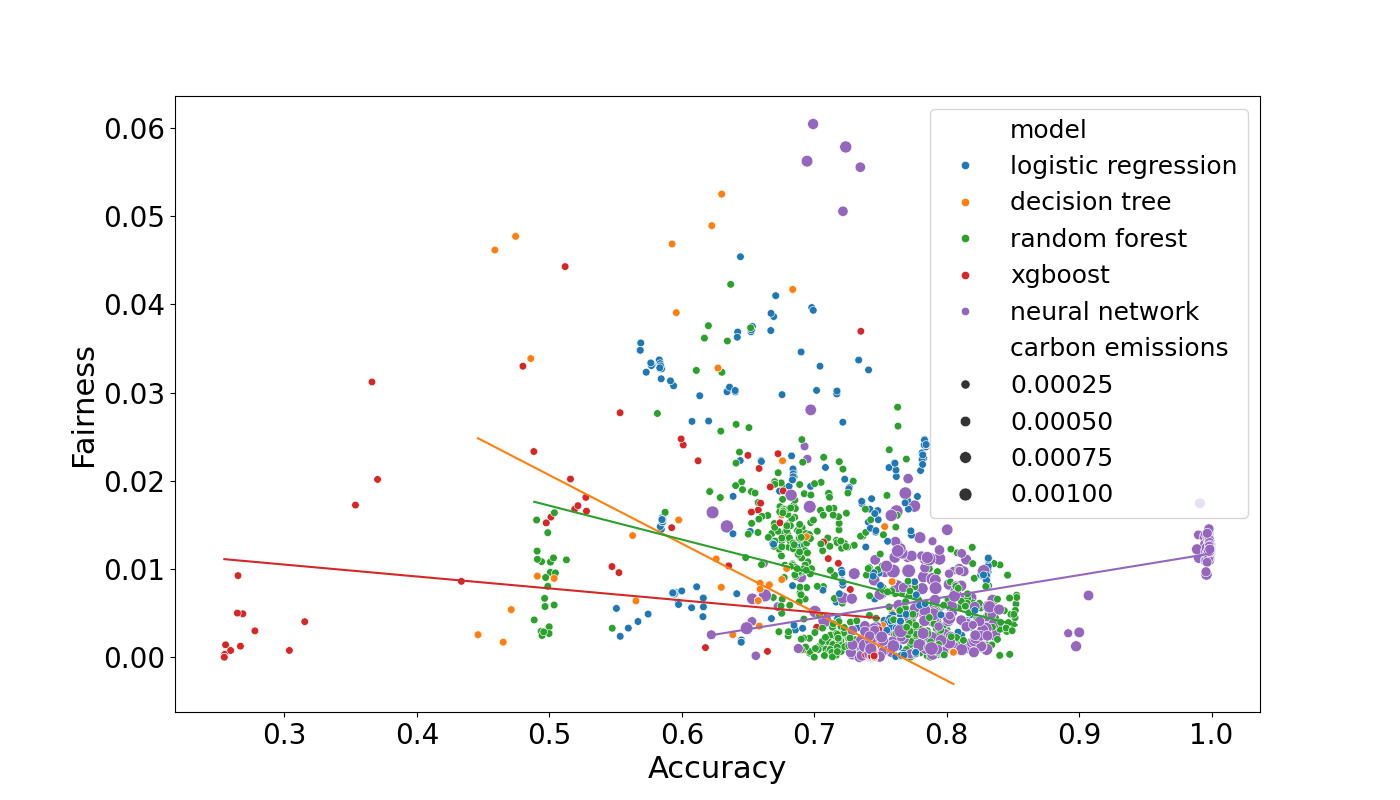}
        \vspace{-12pt}\caption{Loan}\vspace{12pt}
    \end{subfigure}
    % Fig (b)
    \begin{subfigure}{0.33\textwidth}
        \vspace{-12pt}\includegraphics[width=\linewidth]{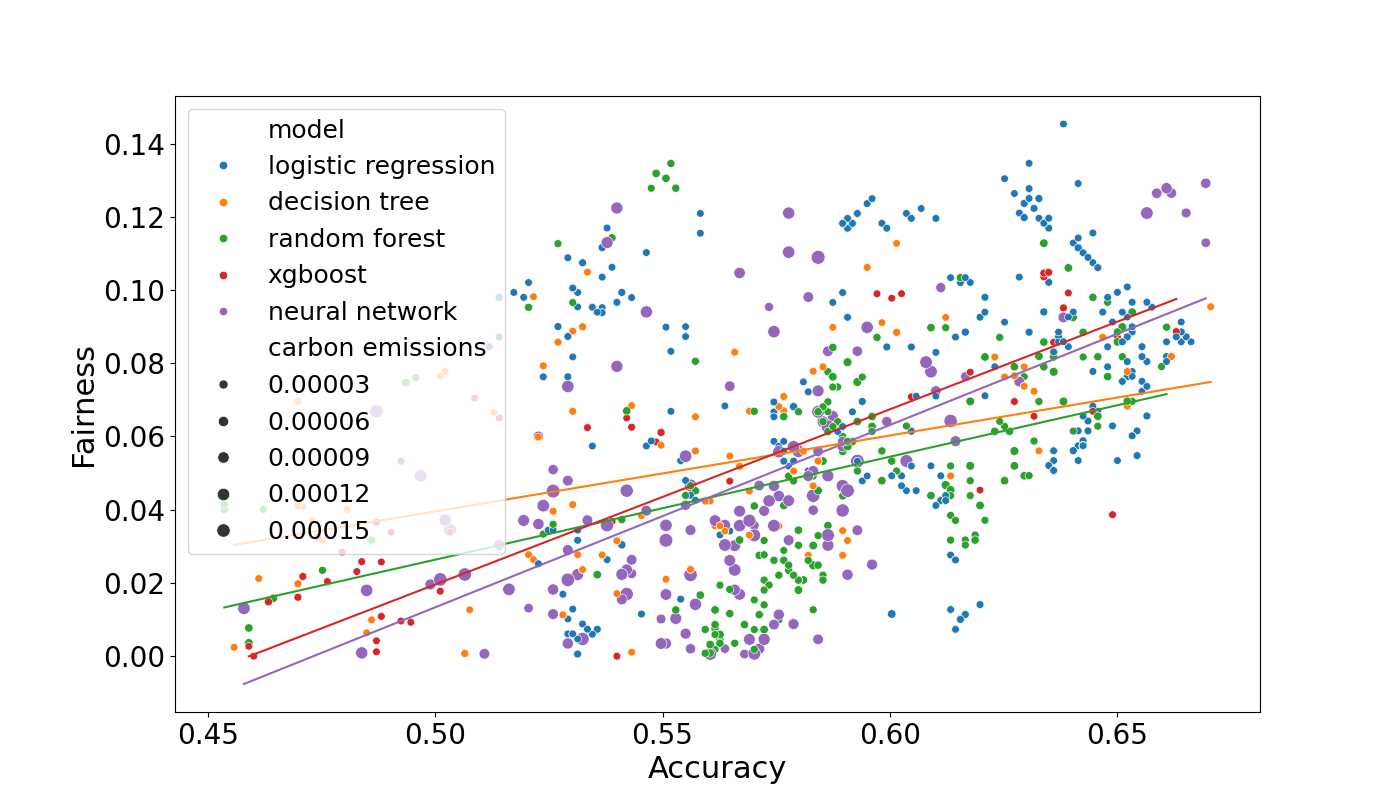}
        \vspace{-12pt}\caption{COMPAS}\vspace{12pt}
    \end{subfigure}
    % Fig (c)
    \begin{subfigure}{0.33\textwidth}
        \vspace{-12pt}\includegraphics[width=\linewidth]{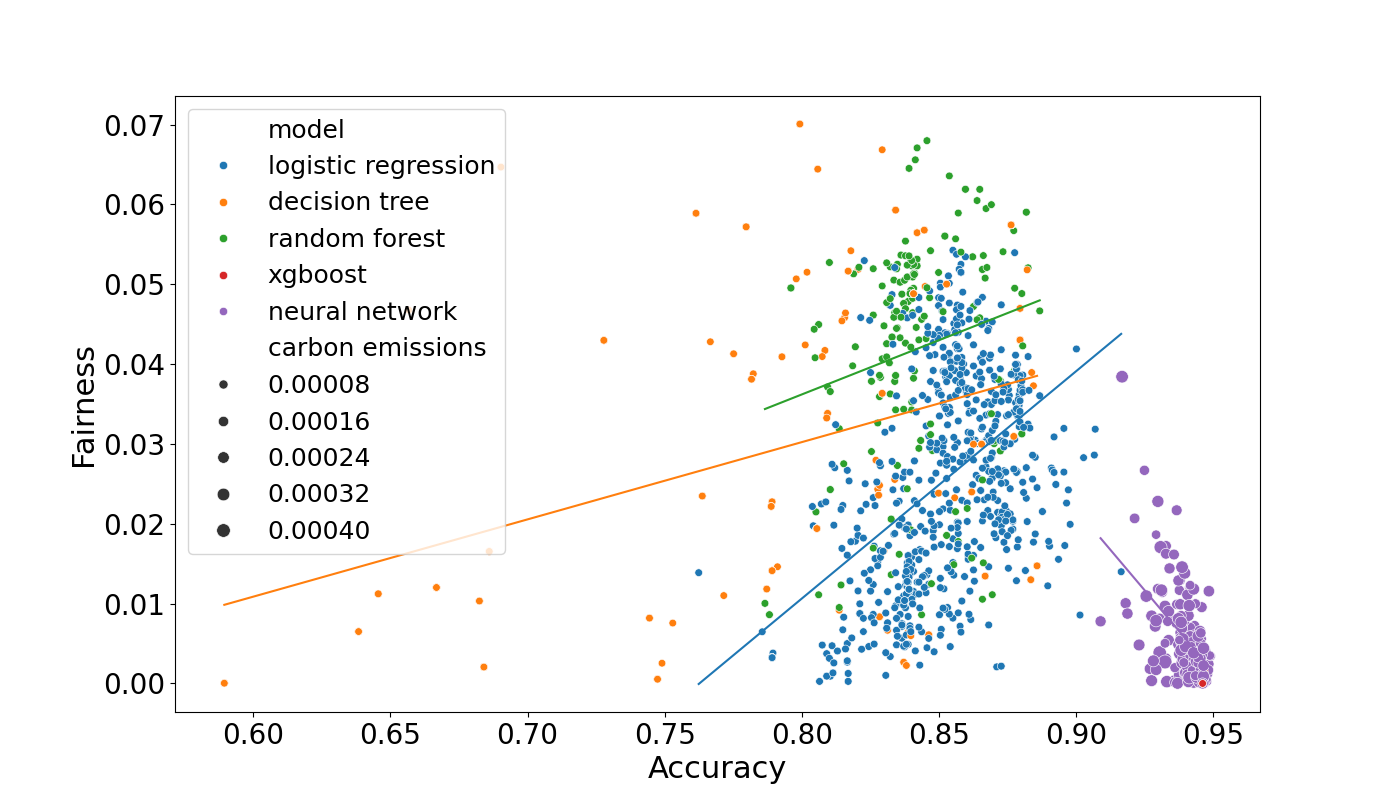}
        \vspace{-12pt}\caption{LSAC}\vspace{12pt}
    \end{subfigure}
    \vspace{-12pt}
\caption{Scatter plots of accuracy against group disparity for different datasets for all $2000$ trials. Different models are shown in different colours, and different levels of carbon emissions are shown using different sizes.}
\vspace{6pt}
\label{fig:optuna_scatter_plots}
\end{figure*}

\subsection{Evaluating GHG Emissions}
We included a GHG emission tracking with CodeCarbon \cite{lottick2019energy} throughout our training and inference pipeline to monitor the carbon emissions from each model training. It helps to track the overall carbon emission by accumulating the power consumption of individual hardware components and converting it into GHG emission based on the energy mixture of local power grids. At the end of the training, the overall GHG emission amount is output along with the model parameters.

\section{Evaluation}
\label{sec:evaluation}

This section presents an experimental analysis of an AI pipeline implementing the proposed FPIG framework. All the code is implemented in Python 3.9, and the experiments are performed on an \textbf{16 CPU} instance consisting of \textbf{64GB} memory. The evaluation is divided into two parts.
Firstly, we run $2000$ trials for each dataset using different models and hyperparameters using the FPIG framework. Using results across all trials, we then identify the relationships between the key metrics (accuracy, fairness, emissions). Secondly, we use the results of the trials to develop a \emph{sustainable meta learning} algorithm, aiming to learn the relationship between the meta-features of the model and dataset used, and the outputted accuracy, fairness and emissions.

\subsection{Dataset, Differential Privacy and Protected Attributes}
\label{subsec:dataset}

We included the five public datasets in our evaluation:
\begin{itemize}
    \item \textbf{Adult Income} \cite{misc_adult_2} is a public multivariate \textit{social} dataset for annual income classification (i.e., if annual income is above 50K). It comprises 48,842 records with 14 attributes such as education, occupation, and work class.
    \item \textbf{COMPAS Recidivism Racial Bias} \cite{compas_2} is a popular commercial algorithm judges and parole officers use to score criminal defendants' recidivism likelihood. This dataset compares the algorithm outputs and the ground truths, which shows that the algorithm is biased in favour of white defendants and against black inmates, based on a 2-year follow-up study.  
    \item \textbf{LSAC} \cite{lsac} is a public dataset originally collected for a 'LSAC National Longitudinal Bar Passage Study' study. It includes background information and if (and how) candidates passed the bar exam to become lawyers in the United States.
    \item \textbf{Loan Default} \cite{loandataset} is a public multivariate \textit{financial} dataset for loan default classification. It includes 139,202 records with 34 attributes such as income, gender, and loan purpose. Note that we randomly sampled 40,000 records from this dataset in our evaluation.
    \item \textbf{Support2} \cite{support2} is a public multivariate \textit{health} dataset for predicting survival over a $180$-day period for seriously ill hospitalized adults. It comprises 9,105 records and 42 attributes, such as age, sex, and follow-up days.
\end{itemize}
Each dataset is split into a training set (70\%) and a test set (30\%). 
We then apply DPView \cite{lin2022dpview} to each training set and generate twenty Differentially Private (DP) synthetic data (having the same number of data points as the training sets) with different DP levels $\epsilon = [0.5, 10.0]$, where smaller $\epsilon$ indicates higher DP. 
Also, we only consider one protected attribute, that is \textit{gender}, for all five datasets. 
We only consider binary protected attributes (thus generating two groups in each case) and measure fairness via disparity, i.e., the absolute difference in the average loss between both groups. This is for simplicity, but our approach can easily carry over to other metrics and more complex settings. 

\subsection{Models and their Parameters}
\label{subsec:model_para}
We built the FPIG framework using various models with varying degrees of complexity. We consider a range of hyperparameters for each model, as detailed below. Varying the model complexity and parameters will give different performance metrics and, better or worse, fairness and GHG emissions. Complex models like Neural Networks are expected to worsen fairness due to overfitting. The trade-offs between fairness and accuracy, as well as GHG emissions, should always be considered when designing new models.
Table \ref{tab:datasets_2} illustrates the models and associated hyperparameters used in our study. We also include a self-defined measure of each model's explainability. This ranges from $1$ to $3$, with $1$ being the most explainable model and $3$ least explainable.

\subsection{Search Space and Pareto-Front Analysis}
\label{subsec:pareto_front}
We exploited Optuna \cite{akiba2019optuna}, a state-of-the-art hyper-parameter optimization package, as our optimization engine to find the Pareto Fronts of the objectives defined in the FPIG framework. For each dataset, we run $2000$ trials. 
For each trial, we select a value between $0.5$ and $10$ for differential privacy, where $10$ indicates lower differential privacy and vice versa. We then use the respective differentially private dataset based on this value. We also include the option of no differential privacy.
The objectives are listed below: 
\begin{itemize}
    \item \textbf{Fairness}: we exploited demographic parity \cite{mehrabi2021survey} as our fairness metrics. The objective is to minimise the group disparity $f$ in Equation \eqref{eq:group_disparity}.
    \item \textbf{Explainability}: ML models are assigned an ordinal value $/{0,1,2,3/}$ to illustrate their explainability as in Table \ref{tab:datasets_2}, the lowest indicating the simplest models. Models range from decision logistic regression to decision trees, random forests, XGBoost, and neural networks.
    \item \textbf{Carbon emission}: CodeCarbon \cite{lottick2019energy} is used to measure GHG emission in practical settings. The GHG emission is reported in the unit of the kilogram. 
    \item \textbf{Accuracy}: given that all the five tasks are classification, we use classification accuracy $[0,1]$ as the performance metric evaluating model utility.
    \item \textbf{Equal Importance}: To find the best models with respect to the accuracy, fairness and carbon emissions, we consider a naive approach for trading off each of them with equal importance. To define this, we scale the values of each metric to be between $0$ and $1$ to obtain $v^D_{i,j}$, the scaled value for each trial $i \in \{1, \cdots, 2000\}$ for each metric $j \in \{\text{Accuracy, Fairness, Carbon Emissions} \}$, for each dataset $D$. The best trial is defined to be 
    \begin{equation}\label{eq:equal_importance}
        i^D = \text{argmin}_{i} \Big( (1-v^D_{i,\text{Accuracy}}) + v^D_{i,\text{Fairness}} + v^D_{i,\text{Carbon Emissions}} \Big),
    \end{equation}
    as we wish to maximise Accuracy, whilst minimising Fairness and Carbon Emissions.
\end{itemize}

\subsection{Optimisation under the FPIG Framework}

We use Optuna to search through the hyperparameter space and find the Pareto-Frontier. Table \ref{tab:best_models} summarises the best models and their overall performances concerning the performance metrics - Accuracy, Fairness, Carbon Emissions - and their corresponding model Explinabilities and Differential Privacy levels. We further illustrate (in a rolling average of 500 trials) the trade-offs between the three performance metrics over time in Figure \ref{fig:optuna_trials}. Below, we summarise our observations.

%%
%% OBSERVATION 1: Trade-off always exist
%%
\paragraph{Although the degree may vary, the trade-offs between objectives always exist.}
%
% Point 1 - Accuracy and Fairness Trade-offs
The results showed that Accuracy can always trade for Fairness across all five datasets. This observation aligns with the heuristic, where better fairness usually leads to worse model utility. However, the degree of trade-offs varies.
As can be seen in Table \ref{tab:best_models}, Adult Income demonstrates a small accuracy reduction from 0.861 to 0.767 when improving the group disparity from 0.167 to 0.0, whilst Support2 shows significant accuracy reductions from 0.977 to 0.255 when improving the group disparity from 0.0259 to 0.0.
%
% Point 2 - Trade of between Carbon Emission and others
Similarly, we can train a model with lower Carbon emissions by trading Accuracy and Fairness. Compared to Accuracy, the trade-offs between Carbon Emission and Fairness are more significant across all five datasets.
For example, when applying the model with the lowest Carbon Emission model to the Adult income dataset, the Accuracy and Group Disparity were reduced by 0.135 and increased by 0.279, respectively, compared to the optimal for Accuracy and Fairness.

%%
%% OBSERVATION 2: Multiple objectives can be improved 
%%
\paragraph{Multiple objectives can be improved during model tuning.}
Although trade-offs (disregarding their degree) are seen across all datasets tested as shown in Table \ref{tab:best_models}, we also observed that multiple objectives could still be improved simultaneously.
%
% Point 1 - Equal Importance
First, the models selected by optimising the Equal Importance (Eq. \eqref{eq:equal_importance}) demonstrate balanced performances between all three metrics - Accuracy, Fairness and Carbon Emission, indicating that it is possible to find a good solution across all datasets.
For example, we find a model in which the Accuracy and Fairness are reduced by only 0.057 and 0.007, respectively, with the COMPAS dataset.
%
% Point 2 - Time series data generates
Second, from Figure \ref{fig:optuna_trials} we observed that the trade-offs between objectives varies as per dataset. For example, Accuracy and Group Disparity are positively correlated (i.e., higher accuracy and lower fairness) in Adult Income and COMPAS datasets, whilst the same correlation in the LSAC and Support 2 datasets is negative. Although it is possible to improve multiple objectives simultaneously, the trajectory toward optimal (concerning the surrogate objective, such as the Equal Importance in Eq. \eqref{eq:equal_importance}) can vary as per dataset.

%%
%% OBSERVATION 3: Fainress lead to lower DP
%%
\paragraph{The models yielding lower Group Disparity are usually more deferentially private.} Heuristically, we know that protected attributes (e.g., gender in our experiments) could potentially be utilised to identify individuals. Therefore, differential privacy (DP) could also be improved when building fair models for those protected attributes. 
Our experiment results shown in Table \ref{tab:best_models} confirm the above hypothesis. The best model concerning fairness always provides better DP (fulfilling smaller privacy budget $\epsilon$) compared to other scenarios.

%% TABLE
%
\begin{table}[t!]
    \centering
    \resizebox{0.50\textwidth}{!}
    {
    \begin{NiceTabular}{|c|c|c|c|c|c|}
    \hline
    \multirow{2}{*}{\textbf{Model Inputs}} & \multicolumn{5}{|c|}{\textbf{Dataset}} \\
    \cline{2-6} & {\textbf{Adult}} & {\textbf{COMPAS}} & {\textbf{Loan}} & {\textbf{LSAC}} & {\textbf{Support2}}
    \tabularnewline \hline \hline
    % Logistic Regression
    Logistic Regression & 0.518 & 0.439 & -0.348 & 0.300 & -0.202
    \tabularnewline \hline
    % Decision Tree
    Decision Tree & 0.472 & 0.383 & -0.608 & 0.253 & -0.321
    \tabularnewline \hline
    % Random Forest
    Random Forest & -0.072 & 0.526 & -0.612 & 0.137 & -0.054
    \tabularnewline \hline
    % XGBoost
    XGBoost & 0.519 & 0.871 & -0.283 & -0.093 & -0.188
    \tabularnewline \hline
    % Neural Network
    Neural Network & 0.485 & 0.449 & 0.493 & -0.570 & 0.028
    \tabularnewline \hline
    \end{NiceTabular}
    }
    \vspace{3pt}
    \caption{The correlation between Accuracy and Group Disparity.} 
    \vspace{.3cm}
    \label{tab:model_performance_corr}
\end{table}

%%
%% OBSERVATION 4: Simpler models are more explainable and GHG friendly. 
%%
\paragraph{Simpler models are usually more explainable and carbon-friendly.}
This observation reinforces our heuristic regarding the trade-offs between model complexity and explainability.
As seen in Table \ref{tab:best_models}, the best models for Carbon Emission adopt the decision tree architecture (i.e., more explainable and easy to compute) across four datasets.
In contrast, Neural Network (NN) and XGBoost, having extensive capability to approximate any continuous function with lower Explainability, achieved the best Accuracy across four datasets. 
Further observations regarding the impact of model architecture will be presented in the next subsection.

% Figure \ref{fig:optuna_trials} shows the rolling average over $500$ trials of the key metrics of interest.
% We note that in the majority of cases, the expected trends previously discussed are followed. For example, in both the Adult Income and Loan Default results, we see that as accuracy increases, we tend to obtain a less fair model (as the group disparity tends also to increase). 
% However, this not always be the case. For example, when looking at the Support2 dataset, we see that we are able to increase accuracy, whilst also decreasing the average group disparity. This is due to the fact that, during optimization, the choice of the next hyper-parameters is stochastic. Hence, there is always a chance of the model selecting non-optimal configurations. 

%%
%
\subsection{The Impact of Model Architecture}

%%
%% Observation 1: Results are dataset dependant (trade-offs and single objective performacne)
%%
The choice of model architecture also plays a significant role when searching for better solutions under the FPIG framework. 
The best model architecture varies significantly across datasets.
We further see this in Figure \ref{fig:optuna_scatter_plots}, where we compare accuracy and fairness for the three datasets - Loan Default, COMPAS and LSAC. We plot a line that best fits each model type and dataset. 
% Trade-offs
As can be seen, the trade-offs not only vary in between datasets but also differ between model architectures. The correlations between Accuracy and Group Disparity when applying different model architectures across the five datasets shown in Table \ref{tab:model_performance_corr} also support this observation. 
In the COMPAS dataset, we see similar behaviour across all model types. As we increase Accuracy, this must come at the expense of Fairness, as Group Disparity also increases. This is shown by the positive gradient of the lines of best fit in Figure \ref{fig:optuna_scatter_plots} and the positive correlations in Table \ref{tab:model_performance_corr}. 
In contrast, the correlation becomes negative when applying NNs to the LSAC dataset, which means that the NN models could improve Accuracy and Group Disparity simultaneously compared to other model architectures.
This difference is less obvious in Loan Default and COMPAS datasets.

%% TABLE - Meta Learning
%%
\begin{table}[t!]
\centering
\begin{tabular}{|c|c|c|c|} 
\hline
 \multirow{2}{*}{\textbf{Model Inputs}} & \multicolumn{3}{|c|}{\textbf{Sustainability Feature Coefficients}} \\
 \cline{2-4}
  & {\textbf{Accuracy}} & {\textbf{Group Disp.}} & {\textbf{GHG}} \\
 \hline
%\textit{Constant} & 12.463 & 0.000 & 0.758 & 0.000 & 0.065 & 0.000 & 0.016 & 0.000 \\ 
\textit{No DP Applied} & 1.118 & 0.843  &  -0.153 \\ 
\textit{Classifier NN (y/n)} & 0.537 & -0.200  & 1.902 \\ 
\textit{DP 5.5 (y/n)} & 0.446 & 0.008  & 0.008 \\ 
\textit{DP 7.0 (y/n)} & 0.402 & 0.215  & -0.018 \\ 
\textit{DP 10.0 (y/n)} & 0.326 & 0.180  & 0.091 \\ 
\textit{Dataset Column Number} & 0.322 & -0.416  & 0.006 \\ 
\textit{DP 3.0 (y/n)} & 0.306 & 0.326  & 0.018 \\ 
\textit{\# of categorical features} & 0.069 & -0.413  & 0.071 \\ 
\textit{\# of NN layers} & -0.024 & 0.021  & 0.417 \\ 
\textit{Size of NN layer} & -0.067 & -0.006  & 0.102 \\ 
\textit{Feature cardinality} & -0.094 & 0.317 & -0.016 \\ 
\textit{Variance of target} & -0.461 & -0.149  & 0.034 \\ 
\hline
\hline
\textit{E-net $R^2$ On Test Set} & 0.6967 & 0.3799 & 0.6806 \\
\hline
\end{tabular}
\vspace{6pt}
\caption{Meta-learning model: Ridge Regression $(\alpha = 1)$ of accuracy, fairness (group disparity) and GHG emissions on selected features of the combined datasets. Coefficients impose the importance of each feature on the metric of interest. Accuracy and fairness are sensitive to differential privacy (lower is more differentially private), whilst GHG emissions depend on the size of the original training set and the model used.}
\vspace{6pt}
\label{tab:automl}
\end{table}

%% SUBSECTION: Meta-Learning
%%
%%
\subsection{Sustainable Meta Learning} \label{sec:sustainable_meta_learning}

In the previous subsection, we studied the trade-offs between sustainable objectives and realised that the preferred hyperparameters and model architectures vary as per dataset. A single solution does not exist that fits all scenarios.
Following the methodology in \cite{yang2019oboe}, we trained regression models $\mathcal{M}_i$ for each of $i \in [ \text{accuracy}, \text{disparity}, \text{emissions}]$ that learn the relationship between the key objectives (i.e. accuracy, group disparity, and GHG emission), based on features of the dataset $d_X$ (e.g., number of features, number of entries) as well as features on the model and training $d_m$ ( e.g., hyperparameters of the model architecture, number of training epochs) and finally privacy level requirements, $d_p$. This is trained based on the trained models from the FPIG framework using Optuna. 
We aim to offer users a framework to determine which architecture and "sustainable hyperparameters" to pick given requirements before training, which is typically time-consuming and leads to extensive energy consumption and GHG emissions.
This is the first step towards developing broader frameworks and attempting to define a meta-learning approach that will allow for a more automated ML system.

%% Meta-Learning Results
%%
Table \ref{tab:automl} summarises the results by running separate regression models against each objective of interest and showing the learnt coefficients of selected inputs. We list the most important features in the Table. Our experiment
demonstrates the relationship between dataset, model hyperparameters and sustainability features.
We observe that using a less differentially private dataset increases the accuracy, thus illustrating the trade-off between accuracy and privacy previously discussed. We also note that using neural networks tends to increase accuracy, whilst using a dataset with a large variance in the target will decrease the model's accuracy. 
Further, group disparity tends to decrease with more privacy. Using no differential privacy is the biggest factor in having a larger group disparity, with a coefficient of 0.843. 
Lastly, we note that using Neural Networks has the largest impact on GHG Emissions, thus increasing with model complexity. There doesn't seem to be any significant relationship between privacy requirements and carbon emissions. 

%Table \ref{tab:automl_2} shows the performance of models with different levels of regularisation and compares it with simple benchmarks. We see that simple linear models can have significant predictive power about accuracy, fairness and emissions. The results show that there is more predictive power in relation to accuracy and emissions, whilst the estimation of group disparity is less determined compared to the others. Further, we note that adding some regularisation tends to increase performance slightly. 

Algorithm \ref{alg:candidate_model_eval} demonstrates how the meta learning algorithms can be used in practice. This takes in a dataset $X$, a set of candidate model architectures $d^{(k)}_m$ for $k = 1, \cdots, K$, along with user requirements $\tau$ on the minimum accuracy, maximum disparity and maximum emissions the user is willing to accept. The algorithm then returns only the model architectures whose estimated metrics sit within the thresholds, based on the meta learning models. From this, a user can select their chosen model based on preference across the metrics.

\begin{algorithm}[t!]
\begin{algorithmic}[H]
\REQUIRE Dataset $X$, meta-learning models $\mathcal{M}_i$ for $i \in [ \text{accuracy}, \text{disparity}, \text{emissions}]$,  user requirements $\tau = [\tau_{\text{acc}}, \tau_{\text{disp}}, \tau_{\text{em}}]$, privacy requirement $d_p$, candidate model architectures $d^{(k)}_m$ for $k = 1, \cdots, K$.
\STATE{Compute $d_x$ for dataset $X$}
\FOR{$k = 1, \cdots, K$}
\STATE{Compute 
\begin{eqnarray*}
    m^{(k)}_\text{acc} & = &\mathcal{M}_\text{accuracy}[ d_x, d_p, d^{(k)}_m ]  \\
m^{(k)}_\text{disp} & = & \mathcal{M}_\text{disparity}[ d_x, d_p, d^{(k)}_m ]  \\
m^{(k)}_\text{em} & = &\mathcal{M}_\text{emissions}[ d_x, d_p, d^{(k)}_m ] 
\end{eqnarray*}
}
\ENDFOR
\RETURN $\{ d^{(k)}_m$: $m^{(k)}_\text{acc} \geq \tau_{\text{acc}} $, $m^{(k)}_\text{disp} \leq \tau_{\text{disp}} $, 
    $m^{(k)}_\text{em} \leq \tau_{\text{em}} \}$
\end{algorithmic}
\caption{Candidate Model Evaluation}
\label{alg:candidate_model_eval}
\end{algorithm}

\section{Conclusion}
\label{sec:conclusion}
 
This paper introduces the FPIG framework for Sustainable Machine Learning, considering a multi-objective optimisation problem involving accuracy, fairness, privacy, explainability and carbon emissions. We demonstrate this approach on five datasets and show the trade-offs between these sustainable objectives and the possibility of finding models to balance these trade-offs.
We also extended these observations by building a meta-learning approach to predict these key metrics based on the dataset and model characteristics.
The results further validate our above observations and show that fairness, as one of the sustainable objectives, is more data-dependent, meaning that it is more difficult to provide guarantees by selecting suitable model architectures and corresponding hyperparameters.  
%%%%%%%%%%%%%%%%%%%%%%%%%%%%%%%%%%%%%%%%%%%%%%%%%%%%%%%%%%%%%%%%%%%%%%%%

%%% Use this command to include your bibliography file.
%\clearpage
\bibliography{references}

%\newpage
%\appendix
%\input{reviews}
%\newpage
%\input{rebuttal}

\end{document}